# Studies with impossible languages falsify LMs as models of human language

Jeffrey S. Bowers, School of Psychology and Neuroscience, University of Bristol

Jeff Mitchell, School of Engineering and Informatics, University of Sussex

Commentary on Futrell, R., & Mahowald, K. (in press). How linguistics learned to stop worrying and love the language models. Behavioural and Brain Sciences.

## Abstract

According to Futrell and Mahowald **(F&M)**, both infants and language models (LMs) find attested languages easier to learn than "impossible languages" that have unnatural structures. We review the literature and show that LMs often learn attested and many impossible languages equally well. Difficult to learn impossible languages are simply more complex (or random). LMs are missing human inductive biases that support language acquisition.

The speed with which infants learn languages is a core challenge for models of human language acquisition. Chomsky's hypothesis is that humans have an inductive bias, a "Universal Grammar" (UG), that not only constrains the linguistic structures of all attested languages, but also, allows children to learn quickly. By contrast, language models (LMs), such as ChatGPT, lack human-like priors, and consequently, they display just the opposite profile, namely, they learn "impossible languages" that violate UG just as easily as attested ones (Mitchell & Bowers, 2020), and at the same time, require training on many orders of magnitude of more language data compared to children (Bowers, 2025a). According to Chomsky, the ease with which LMs learn impossible languages undermines LMs as theories of human languages. For instance, Chomsky et al., (2023) write:

> "Their deepest flaw is the absence of the most critical capacity of any intelligence: to say not only what is the case… but also what is not the case and what could and could not be the case…
>
> ChatGPT and similar programs are, by design, unlimited in what they can "learn" (which is to say, memorize); they are incapable of distinguishing the possible from the impossible. Unlike humans, for example, who are endowed with a universal grammar that limits the languages we can learn to those with a certain kind of almost mathematical elegance, these programs learn humanly possible and humanly impossible languages <u>with equal facility</u>" [the underlined text encoded a link to Mitchell and Bowers, 2020].

Since Mitchell and Bowers (2020) there have been several studies that have compared how easily LMs learn attested and impossible languages. For instance, Kallini et al. (2024) reported two impossible languages that were more difficult to learn than English and concluded that LMs (and humans) do not need any linguistic priors to account for the possible-impossible learning gap. **F&M** describe this study and cite several others (including Mitchell and Bowers) in a way that suggests that LMs consistently show a human-like difficulty in learning impossible languages, writing:

> "Kallini et al. (2024) find that the model learns from real English text consistently faster than these baselines (see also Mitchell and Bowers, 2020; Yang et al., 2025; Xu et al., 2025; Ziv et al., 2025, among others). These results and others show that Transformers and related models have inductive biases that align with human language".

But this characterisation of the research is mistaken. First consider the Kallini et al study. Although the authors reported two impossible languages that were more difficult to learn, they reported several additional impossible languages were learned almost as easily as English, including one of the impossible languages that Mitchell and Bowers (2020) devised. Furthermore, the impossible language that was the most difficult to learn was composed of random shuffles of words. That is, there was no structure to learn. The other difficult-to-learn impossible language was composed of deterministic random shuffles, with sentences of different lengths shuffled in different deterministic ways (e.g., all sequences of length 5 were shuffled in one order, sequence of length 6 shuffled in another order, etc.). In this case, there was something to learn, namely, different random languages for the different sentence lengths. The observation that it is more difficult to learn multiple compared to a single random language does not challenge Chomsky's point. More generally, a learning algorithm that performs well on one data type must perform badly on others ("no free lunch theorems"; Wolpert & Macready, 2002). Accordingly, the observation that LMs perform badly on some languages is an unnecessary empirical confirmation of these theorems. In contrast, the success in learning some impossible languages is a falsification of the claim that LMs are an adequate model of human language learning.

The same issue applies to Yang et al. (2025) article that **F&M** cited. The authors again assessed LMs on a range of impossible languages, including the same deterministic shuffled languages used by Kallini et al. And again, the authors found that several impossible languages were easily learned, whereas the random shuffle language were difficult. And again, the slow learning of the random shuffled language was taken as problematic for Chomsky:

> "If LMs function as non-human like pattern recognizers as Chomsky et al. (2023); Moro et al. (2023) argue, they should be able to learn these languages as well as attested ones".

But again, the observation that it is more difficult to learn multiple different impossible languages (for different sentence lengths) is not surprising on any theory.

The Xu et al. (2025) article that **F&M** cited varied the plausibility of languages (rather than the impossibility of languages). In this case, the authors found that LMs struggled with some implausible languages, but again, in some cases, the LM found the implausible languages easy to learn. And with regards to conditions in which LMs struggled, the authors note that there may have been errors in the construction of the materials that weaken the conclusions that can be drawn:

> "Thus, it is possible that our findings may be due partially or entirely to increased noise in the counterfactual corpora, rather than inherent differences in learnability between the original and counterfactual grammars".

It is difficult to take these findings as problematic for Chomsky's thesis.

Finally, the Ziv et al. (2025) article that **F&M** cite reports several impossible languages that LMs learn just fine, including partially reversed languages (replicating Mitchell and Bowers, 2020). In addition, in another study not cited, Lou et al. (2024) report that LMs can happily learn reversed languages.

So, contrary to the claims of **F&M**, LMs often learn impossible languages as easily as attested languages. At same time, LMs require many orders of magnitude more training than infants (e.g., Bowers, 2025a), and a recent attempt to induce a universal grammar in LMs (McCoy & Griffiths, 2025) does not make LMs much more data efficient (Bowers, 2025b). Given all this, the appropriate conclusion is that current LMs learn in just the way one would expect of models that are missing human-like innate biases for language learning.